# INCREMENTAL MAINTENANCE OF ASSOCIATION RULES UNDER SUPPORT THRESHOLD CHANGE

Mohamed Anis BACH TOBJI
*BESTMOD Laboratory – Institut Supérieur de Gestion*
*41, rue de la liberté, cite Bouchoucha*
*Bardo, 2000, Tunis*
anis.bach@isg.rnu.tn

Mohamed Salah GOUIDER
*BESTMOD Laboratory – Institut Supérieur de Gestion*
*41, rue de la liberté, cite Bouchoucha*
*Bardo, 2000, Tunis*
ms.gouider@isg.rnu.tn

**ABSTRACT**

Maintenance of association rules is an interesting problem. Several incremental maintenance algorithms were proposed since the work of (Cheung et al, 1996). The majority of these algorithms maintain rule bases assuming that support threshold doesn't change. In this paper, we present incremental maintenance algorithm under support threshold change. This solution allows user to maintain its rule base under any support threshold.

## 1. INTRODUCTION

The problem of association rules mining (ARM) (Agrawal et al, 1993) receives a lot of attention. This attention is motivated by several application domains such as market basket analysis, telecommunications analysis, Web logs analysis etc. Several ARM algorithms have been constructed to solve this problem (Agrawal and Srikant, 1994), (Savasere et al, 1995), (Zaki et al, 1997), (Pasquier et al, 1999), (Han et al, 2000). They compute association rules that describe a data set. Nevertheless, data sources are frequently updated. It follows that extracted rules describe a data set at a moment $t$, the mining moment. This change of data implies invalidation of rules and necessity to maintain them.

Several algorithms of incremental maintenance have been proposed (Cheung et al, 1996), (Tsai et al, 1999), (Zhang et al, 1997); the first algorithm called FUP (Cheung et al, 1996) was achieved in 1996. These algorithms target maintenance of rule bases under the same support threshold. In this paper, we present IMSC, an algorithm of incremental maintenance after several updates (insertion of transactions). IMSC allows the user to change support threshold. This algorithm is based on Apriori (Agrawal and Srikant, 1994) that is oriented sparse databases such as transactional databases.

Generally, maintenance of association rules is achieved according to one of the two following approaches; incremental approach and non-incremental approach. Non-incremental maintenance consists in executing again an ARM algorithm on the whole updated data. The advantage of this method is the free choice of a new support threshold. Its drawback is ignoring old association rules making the operation more expensive. Incremental maintenance is the second approach. This method uses old association rules to compute new association rules. It's faster in comparison with non-incremental maintenance. Nevertheless, its drawback is that maintenance is realized on the same initial extraction support threshold. The purpose of this work is to develop an incremental maintenance algorithm under support threshold change. Our algorithm combines the advantages of the two maintenance approaches.

The remaining of the paper is organized as follows: in section 2, we define the problem of association rule mining and maintenance, and in section 3 we present our algorithm. The performance of the algorithm is studied in section 4, and finally, conclusion and perspectives are given in section 5.

## 2. PROBLEM DESCRIPTION

Let $I$ be a set of $n$ items, $I= \{i_1, i_2, i_3,...., i_n\}$. Let $BD$ be a database of $D$ transactions with schema *<tid,items>*. Each transaction is included in $I$. An association rule is $X \rightarrow Y$ with $X,Y \subset I$, $X \neq \emptyset$ and $X \cap Y = \emptyset$. Rule *support* is the occurrence number of $X \cup Y$ in $BD$, and its *confidence* is $support(X \cup Y)/support(X)$. Given a minimum support threshold s% and a minimum confidence threshold c%, ARM problem (Agrawal et al, 1993) is to find out all association rules which support exceeds s% and confidence exceeds c%. An itemset is a set of items. It said to be frequent in $BD$ if its support in $BD$ ($X.support_{BD}$) exceeds s% ($X.support_{BD} \geq s \times D$), we say that is *s-frequent* in $BD$. The ARM problem may be reduced to the computation of frequent itemsets, because once we have frequent itemsets set, association rules generation will be straightforward.

Let be $F$ the set of the frequent itemsets in $BD$ and $s$ the support threshold under which $F$ is extracted. After several updates of $BD$, an increment $bd$ of $d$ transactions is added to $BD$. The problem of incremental maintenance under support threshold change is to compute $F'$, the set of the frequent itemsets in $BD'=BD \cup bd$ according to a support threshold $s'$.

## 3. THE IMSC APPROACH

### 3.1 Winner, loser and persistent itemsets

To solve this problem, we propose a new approach we call IMSC (Incremental Maintenance of association rules under Support threshold Change) that is based on FUP (Fast UPdate) algorithm (Cheung et al, 1996). FUP maintains a rule base incrementally under the same support threshold. When s=s', IMSC and FUP are practically identical.

Let be $BD'=BD \cup bd$. $BD'$ contains four types of itemsets:

- *Winner* itemsets that are not s-frequent in $BD$ ($\notin F$) and that are s'-frequent in $BD'$ ($\in F'$).
- *Persistent* itemsets that are s-frequent in $BD$ ($\in F$) and that are s'-frequent in $BD'$ ($\in F'$).
- *Loser* itemsets that are s-frequent in BD ($\in F$) and that are not s'-frequent in $BD'$ ($\notin F'$).
- The itemsets that are not s-frequent in BD ($\notin F$) and that are not s'-frequent in $BD'$ ($\notin F'$).

We are interested in winner, persistent and loser itemsets.

**Definition 1 (persistent itemset)**

*An itemset X is persistent in BD' if $X \in F$ and $X \in F' \Rightarrow X.support_{BD} \geq s \times D$ AND $X.support_{BD'} \geq s' \times (D+d)$.*

**Definition 2 (winner itemset)**

*An itemset X is winner in BD' if $X \notin F$ and $X \in F' \Rightarrow X.support_{BD} < s \times D$ AND $X.support_{BD'} \geq s' \times (D+d)$.*

**Definition 3 (loser itemset)**

*An itemset X is loser in BD' if $X \in F$ and $X \notin F' \Rightarrow X.support_{BD} \geq s \times D$ AND $X.support_{BD'} < s' \times (D+d)$.*

## 3.2 Description of IMSC

The IMSC algorithm browses search space (itemsets lattice) in breadth-first way; it computes frequent itemsets level by level. In other words, it computes frequent itemsets of size $i$ before computing itemsets of size $i+1$.

For each lattice level, frequent itemsets computation is done in two steps. Let's suppose we want to generate the set $F'_i$ (set of s'-frequent itemsets of size $i$), IMSC proceeds as follows:

**I – IMSC scans *bd* :**

1- Generation of $C_i$, set of candidate itemsets of size $i$. $C_i = Apriori\_gen(F_{i-1})/F_i$. Apriori_gen being Apriori candidate generation function defined in (Agrawal and Srikant, 1994). We eliminate the set $F_i$ because it will be treated in next phase (**II**).

2- While scanning $bd$, we update itemsets supports of $F_i$ and $C_i$.

3- We insert in $F'_i$ persistent itemsets which are generated from $F_i$ ($X.support_{BD'} \geq s' \times (D+d)$).

4- The set $C_i$ contains possible winners. A candidate itemset is a possible winner if its support in $bd$ exceeds the *Candidate Pruning Threshold* noted *CPT* (see lemma 1).

**Lemma 1 (Candidate Pruning Threshold)**

*A candidate itemset X is a possible winner in BD' if and only if its support in bd exceeds the threshold CPT, $CPT = s' \times d + (s'-s) \times D + 1$.*

**Proof**

*X is a possible winner if :*

$X.support_{BD'} \geq s' \times (d+D) \Rightarrow X.support_{bd} + X.support_{BD} \geq s' \times (d+D) \Rightarrow X.support_{bd} \geq s' \times (d+D) - X.support_{BD}$

*We know $X \notin F$, so $X.support_{BD} < s \times D$, the maximum support value of X in BD is ($s \times D - 1$).*

$\Rightarrow X.support_{bd} \geq s' \times (d+D) - s \times D + 1 \Rightarrow X.support_{bd} \geq s' \times d + D \times (s'-s) + 1$.

**II– IMSC scans *BD*:**

1- Updating candidate itemsets supports. At the end we have their supports in BD'.

2- Insertion of winner itemsets in $F'_i$.

IMSC iterates $n+1$ times to compute $F'$, where $n$ is the size of the longest s'-frequent itemset. Such as FUP, the return to $BD$ is achieved to generate the winner itemsets from candidate itemsets set. The pruning of this set is made in step **I.4** according to the threshold *CPT*. But *CPT* can be negative (in this case, no candidate itemset is discarded), or superior to $d$ (in this case, all itemsets candidates are discarded). We distinguish three possible scenarios: $CPT \leq 0$, $CPT > d$, $0 < CPT \leq d$.

- **CPT ≤ 0**

In this case, each candidate itemset $X$ satisfies the $X.support_{bd} \geq CPT$ condition, we do not eliminate any candidate. The return to $BD$ will be heavy, because it's about computing all candidate itemset supports. We also know that the s-frequent itemsets in $BD$ (the set $F$) are s'-frequent in $BD'$ (belong to $F'$) thanks to the following lemma.

**Lemma 2**

*In the case CPT≤0, there is no possible losers, all s-frequent itemsets in BD are s'-frequent in BD'.*

**Proof**

*Let X be an itemset that is s-frequent in BD and CPT ≤ 0.*

$CPT \leq 0 \Rightarrow s' \times d + (s'-s) \times D + 1 \leq 0 \Rightarrow s' \times (d+D) - s \times D + 1 \leq 0$

$\Rightarrow s \times D - 1 \geq s' \times (d+D)$ ; we know that $X.support_{BD} \geq s \times D$

⇒ *X.support*$_{BD}$ ≥ *s'×(d+D)* ; we know that *X.support*$_{BD'}$ ≥ *X.support*$_{BD}$ ⇒ *X.support*$_{BD'}$ ≥ *s'×(d+D)*

⇒ *X is s'-frequent in BD'*.

- **CPT > d**

In this case, it is useless to generate candidate itemsets set since no itemset can win (it needs more than *d* occurrences in *bd* ; it's impossible). There is no return to *BD* (step **II**) and the scan of *bd* is sufficient to calculate *F'*, this case is the best one.

- **0 < CPT ≤ d**

In this case, candidate itemsets can win, the pruning is made according to *CPT* that is a moderate value.

## 3.3 IMSC algorithm

In this section, we present the IMSC algorithm which is composed of three procedures; each procedure corresponds to one scenario. Algorithm description is given in pseudo code commentaries:

**IMSC Algorithm :**

Input :

`BD` : Initial database.

`D` : Cardinal of BD.

`bd` : data increment.

`d` : Cardinal of bd.

`F=` ∪ `Fk` (Union of sets Fk).

`s` : support under which F is computed.

`s'` : support under which we want maintain F.

Output:

`F'=` ∪`F'k` (Union of sets F'k).

Algorithm:

```
Begin
01 cpt=s'×d/100+D×(s'-s)/100+1 /*CPT
computation from algorithm parameters and execution
of the convenient procedure*/
02 If cpt ≤ d and cpt > 0 then
03    IMSC1(BD, D, bd, d, F)
04 ElseIf cpt < 1 Then
05    IMSC2(BD, D, bd, d, F)
06 Else
07    IMSC3(BD, D, bd, d, F)
08 EndIf
End
```

**IMSC1 Procedure**

Same input and output of IMSC.

```
Begin
01 F[1].CopyIn(Fk) /*Fk contains size one s-
frequent itemsets in BD*/
02 Scanbd (Fk, Ck) /*While scanning bd, supports
of F₁ are updated and C₁ is created*/
03 min_supp=s'×(d+D)/100
04 Fk.Prune (min_supp) /*Items of F₁ that are not
s'-frequents in BD' are discarded*/
05 cpt=s'×d/100 + D×(s'-s)/100
06 Ck.Prune(cpt) /*Items of C₁ which supports
doesn't exceed cpt are discarded*/
07 ScanBD(Ck) /* While scaning BD, supports of C₁
are updated*/
08 Ck.Prune (min_supp) /*Only items that are s'-
frequent in BD' are kept in C₁*/
09 F'k=Union(Fk,Ck) /*F'₁ is the union of F₁ and
C₁*/
10 While (F'k.Empty = false) Do /*F'
Computation is done level by level*/
11    F'.Insert(F'k) /*F'k is inserted in F'*/
12    k=k+1 /*Going next level*/
13    F[k].CopyIn(Fk)
14    Apriori_gen(Ck,F'k) /*Making Ck that is
Aprioi_gen of F'k*/
15    Fk=Fk.Intersect(Ck) /*Only Itemsets
belonging to Fk∩Ck can be persistent, they need only
to a bd scan because we know their support in BD*/
16    Ck=Ck.Minus(Fk) /*Other itemsets of Ck need a
BD' scan because we don't know their support in BD*/
17    Scanbd(Fk, Ck) /*Computing itemsets support
in bd for Fk and Ck*/
18    Fk.Prune(min_supp) /*We discard infrequent
itemsets in Fk, we obtain persistent ones*/
19    Ck.Prune(cpt) /*We discard itemsets in Ck
that cannot win*/
20    ScanBD(Ck) /*Computation of the remaining
itemsets support of Ck in BD*/
```

```
21   Ck.Prune(min_supp) /*Computation of
winner itemsets*/
22   F'k=Union(Fk,Ck) /*F'k contains persistent
and winner itemsets*/
23 EndWhile
End
```

**IMSC2 Procedure**

Same input and output of IMSC.
```
Begin
01 F[1].CopyIn(Fk)
02 Scanbd(Fk, Ck)
03 min_supp=s'×(d+D) / 100
04 Fk.Prune(min_supp)
05 ScanBD(Ck,F[1])  /* Items not belonging to C1
and F[1] are inserted in C1 ; they can be s'-frequent in
BD' */
06 Ck.Prune(min_supp)
07 F'k=Union(Fk,Ck)
08 While (F'k.Empty = false) Do
09   F'.Insert(F'k)
10   k=k+1
11   F[k].CopyIn(Fk)
12   Apriori_gen(Ck,F'k)
13   Ck=Ck.Minus(Fk)
14   Scanbd(Fk, Ck)
15   ScanBD(Ck)
16   Ck.Prune(min_supp)
17   F'k=Union(Fk,Ck)
18 EndWhile
End
```

**IMSC3 Procedure**

Same input and output of IMSC.
```
Begin
01 F[1].CopyIn(Fk)
02 Scanbd(Fk)
03 min_supp=s'×(d+D) / 100
04 Fk.Prune(min_supp)
05 F'k=Union(Fk,Ck)
06 While (F'k.Empty = false) Do
07   F'.Insert(F'k)
08   k=k+1
09   F[k].CopyIn(Fk)
10   Scanbd(Fk)
11   Fk.Prune(min_supp)
12   F'k=Union(Fk,Ck)
13 EndWhile
End
```

## 3.4 IMSC example

The IMSC algorithm process three possible scenarios to maintain the set of frequent patterns under support threshold change. Let be an initial database *BD* (see figure 1), an increment *bd* and different values of *s* and *s'*. In the following, we execute IMSC to maintain *F*, i.e., to compute *F'*:

- **First scenario : 0 < CPT ≤ d**

Let s=30%, then F is {$A_7$, $B_5$, $C_6$, $D_4$, $AB_4$, $AC_4$, $BC_4$, $CD_3$, $ABC_3$}.

Let *bd* be an increment of data containing the following transactions (each transaction is a doublet (Tid, Items)): {(11,ABD),(12,BD),(13,BCD)}, we have d=3.

Let s' =35%, candidate pruning threshold (*cpt*) is 2,55, it is the first scenario. We compute $C_1$ and $F_1$. $F_1$={$A_8$, $B_8$, $C_7$, $D_7$} and $C_1$=Ø. All items of $F_1$ are inserted in $F'_1$ because their supports in *BD'* exceeds 0,35×13=4,55. $C_2$ is calculated from $F'_1$ via Apriori_gen function, we remove elements of $F_2$ from $C_2$: $C_2$={AD,BD}. After a scan on *bd*, we get the supports in *bd* of candidate itemsets of size 2 ; $C_2$={$AD_1$,$BD_3$} and the supports of frequent itemsets in *BD'* ; $F_2$={$AB_5$, $AC_4$, $BC_5$, $CD_4$}. We prune $C_2$ and $F_2$ according respectively to the thresholds 2,55 (*cpt*) and 4,55 (min_supp). $C_2$={$BD_3$}, $F_2$={$AB_5$, $BC_5$}. Elements of $F_2$ are added to $F'_2$. Scan on *BD* is made to update the support of the itemset "BD". Its support is 5, the itemset "BD" is inserted in $F'_2$. $F'_2$ = {$AB_5$, $BC_5$,$BD_5$}. We go to level 3, $C_3$=Ø, $F_3$={$ABC_3$}. While scanning *bd*, support of itemset "ABC" doesn't change, it's rejected, $F'_3$ will be empty and the algorithm stops.

Figure 1. Database example *BD* and its corresponding itemset lattice

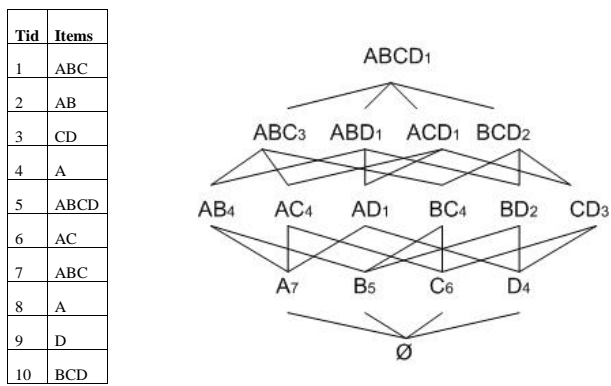

- **Second scenario : CPT ≤ 0**

    Let s=50%, then F is {$A_7$, $B_5$, $C_6$}.

    Let *bd* be a data increment of size *d*=3 containing the following transactions {(11,AB),(12,BC),(13,C)}.

    Let be s' =25%, *CPT* is equal to -0,75. In this case, we are certain there is no loser itemsets; F is included in F' (see lemma 2). After a scan of *bd*, we get $C_1$=Ø, $F_1$={$A_8$, $B_7$, $C_8$}. Elements of $F_1$ are inserted in $F'_1$. Then a scan is achieved on *BD*. During this scan the items that are not in $F_1$ and not in $C_1$ are inserted in $C_1$ because a non frequent item in *BD* according to s could be frequent in *BD'* according to s'. $C_1$={$D_4$}. After pruning $C_1$ according to support threshold 0,25×13=3,25, the only item of $C_1$ is added to $F'_1$, $F'_1$={$A_8$, $B_7$, $C_8$, $D_4$}. $C_2$ is generated from $F'_1$; $C_2$=Apriori_gen($F'_1$)-$F_2$={AB, AC, AD, BC, BD, CD}. $F_2$ being empty, the scan on *bd* is made to compute the supports of the elements of $C_2$, then a scan on *BD* is achieved to update the supports of all itemsets of $C_2$. $C_2$={$AB_5$, $AC_4$, $AD_1$, $BC_5$, $BD_2$, $CD_3$}. After pruning $C_2$ according to s'×(d+D)=3,25, we get $C_2$={$AB_5$, $AC_4$, $BC_5$}, the elements of $C_2$ are inserted in $F'_2$. Then, $C_3$ is generated from $F'_2$, we obtain $C_3$={ABC}. After scanning *bd* and *BD*, the support of "ABC" in *BD'* is 3, $F'_3$=Ø and the algorithm stops.

- **Third scenario : CPT > d**

    Let s=30%, F is then {$A_7$, $B_5$, $C_6$, $D_4$, $AB_4$, $AC_4$, $BC_4$, $CD_3$, $ABC_3$}.

    Let *bd* be a data increment containing the following transactions {(11,AB),(12,BC),(13,C)}, we have *d=3*.

    Let s'=40%, *CPT* is equal to 3,2 which is superior to *d=3*. In this case, there are no candidate itemsets. We must update the supports of elements of F, and to filter them. $F'_1$={$A_8$, $B_7$, $C_8$}. We go to second level since $F'_1$ is not empty. $F_2$={$AB_4$, $AC_4$, $BC_4$, $CD_3$} is updated while scanning *bd*, we have $F_2$={$AB_5$, $AC_4$, $BC_5$, $CD_3$}. The pruning of $F_2$ according to support threshold (=5,2) eliminates all its itemsets. $F'_2$ being empty, the algorithm stops.

## 4. PERFORMANCE STUDY

In this section, we present experimentations achieved to test IMSC efficiency. We implemented IMSC and Apriori and we tested them on synthetic dataset T5.I2.D100k. Generation of synthetic dataset follows the same technique as (Agrawal et al, 1993).

In the first experimentation, we divided database in two partitions: BD that is constituted of 70K transactions and bd that is constituted of 30K transactions. The initial extraction is done on BD according to a support threshold of 30%. The set of the resulting frequent itemsets of this extraction is stored in F. Having BD, bd, s% and F, we executed IMSC to compute F' under several maintenance support threshold going from 1% to 60%. In the same way, we executed Apriori on BD' (= BD∪bd) under the same support threshold interval and we placed the two algorithms curves on the same graphic. The result is shown in the following figure.

Figure 2. Maintenance of frequent itemsets by Apriori and IMSC

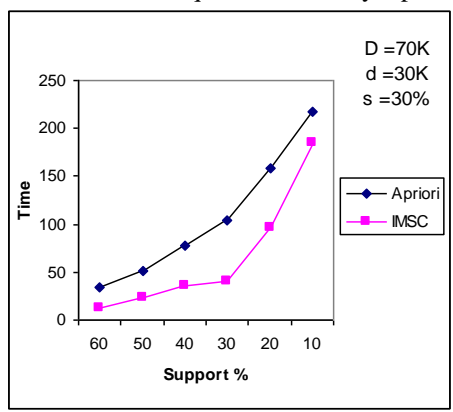

It is obvious that IMSC is faster than Apriori. Two reasons explain this fact. First, IMSC has in memory the initially frequent itemsets in *BD* and their supports. It needs to scan only *bd* to compute their supports in *BD'*. In contrast, Apriori scans all *BD'* to compute their supports. Secondly, Apriori generates the candidates set $C_k$ from $F'_k$ (Apriori_gen), whereas IMSC calculates $C_k$ in the same way except that it eliminates the former frequent itemsets of it ($C_k=C_k-F_k$) and that it prunes it ($C_k$) according to *CPT* (Candidate Pruning threshold). This double pruning of $C_k$ is done just after scaning the increment *bd* that is generally smaller than initial database *BD*.

We also tested IMSC performance for different increment sizes. We executed IMSC for two increments: D=70k, d=30k and s=30%, and D=70k, d=10k and s=30%. The following figure shows experimentation result. It is obvious that smaller increment is, faster IMSC is.

The two vertical lines of every curve delimit the three possible scenarios of our algorithm. For each curve, supports between the two lines make that CPT is between 0 and d. The supports on the left make CPT superior to d, and on the right those that make CPT lower to 0. We notice that when CPT is superior to d, the execution time is nearly null. When CPT is between 0 and d, more support increases, more the curve accelerates (the execution time accelerates), it is due to the return to the initial database. But this acceleration accentuates more in the last case, because of the heavier return to initial database.

Figure 3. IMSC Test under increment variation

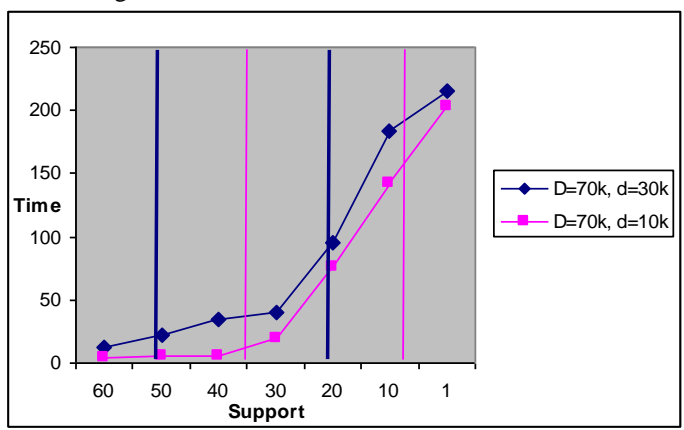

We implemented our IMSC algorithm and we tested it against an ARM algorithm (Apriori). The results we obtained were expected; the incremental maintenance achieved by IMSC is faster than a new frequent itemsets extraction by an ARM algorithm, and IMSC curves have the same shape in which we distinguish the three maintenance scenarios.

## 5. CONCLUSION AND PERSPECTIVES

In this paper, we propose an incremental maintenance approach for the maintenance of association rule base after data source updating under support threshold change. This approach (IMSC) is Apriori like. Apriori is efficient for mining sparse databases (such as transactional databases). Many other ARM algorithms are more efficient when it is about correlated database (Pasquier et al, 1999), (Bastide et al, 2002). It's interesting to develop methods for incremental maintenance under support threshold change based on these algorithms for correlated databases.

We also propose to integrate the incremental maintenance idea in inductive databases (Imielinski and Mannila, 1995), i.e., incremental pattern maintenance under several types of constraints and not only the anti-monotone constraints such as for IMSC.

Many algorithms use database vertical representation (Eclat (Zaki et al, 1997), Partition (Savasere et al, 1995) etc.); Support computation of candidate itemsets is done via tidlists intersection. These algorithms are deficient when databases are very large because of their memory need. However, in the maintenance problem, increment size is small. We think that using this strategy for itemset support computation should improve IMSC.